# A new band selection approach based on information theory and support vector machine for hyperspectral images reduction and classification

A.Elmaizi, E.Sarhrouni, A.hammouch, C.Nacir
Electrical Engineering Research Laboratory, ENSET Mohammed V University
Rabat, Morocco
asma.elmaizi@gmail.com.   sarhrouni436@yahoo.fr.   hammouch_a@yahoo.com.   nacir_chafik@yahoo.fr

*Abstract*—The high dimensionality of hyperspectral images consisting of several bands often imposes a big computational challenge for image processing. Therefore, spectral band selection is an essential step for removing the irrelevant, noisy and redundant bands. Consequently, increasing the classification accuracy. However, identification of useful bands from hundreds or even thousands of related bands is a nontrivial task. This paper aims at identifying a small set of highly discriminative bands, for improving computational speed and prediction accuracy. Hence, we proposed a new strategy based on joint mutual information to measure the statistical dependence and correlation between the selected bands and evaluate the relative utility of each one to classification. The proposed filter approach is compared to an effective reproduced filters based on mutual information. Simulations results on the hyperspectral image HSI AVIRIS 92AV3C using the SVM classifier have shown that the effective proposed algorithm outperforms the reproduced filters strategy performance.

## I. INTRODUCTION

Hyperspectral sensors collect the images simultaneously in hundreds of contiguous spectral bands, with wavelengths ranging from visible to infrared [1][2]. It records the subtlest variations in the surface-reflected solar energy allowing for efficient and unequivocal identification of the observed targets. The hyperspectral imaging is used in many areas, including medical domains, security and defense areas, monitoring and target recognition, mining and oil exploration, agriculture fields, and food safety areas.

The technology offers the capacity to observe and discriminate unique characteristics of materials and features, it provides a rich, detailed spectral information that helps detecting targets and classifying materials with potentially higher accuracy.
However, it brings challenges for hyperspectral image processing. In fact, the highly correlated spectral bands have a degree of redundancy, in addition to the noisy irrelevant bands that cost extra computation burden for hyperspectral image processing.

Consequently, dimensionality reduction [4] is crucial for hyperspectral images. The measured bands are not necessarily all needed for an accurate discrimination and the use of the entire set of bands can lead to a poor classification model. This is due to the curse of dimensionality described by Huges [3].
This paper introduces a new joint mutual information-based approach for HSI dimensionality reduction and suggests that the proposed approach can provide a deeper understanding and a better classification results based on the bands relevance, interaction maximization and redundancy minimizing.
The approach was evaluated using HSI AVIRIS 92AV3C [16] provided by the NASA with the support vector machine classifier. The Indian Pines image was gathered by Airborne Visible/Infrared Imaging Spectrometer (AVIRIS) sensor over Northwest Indiana's Indian Pines test site in June 1992.

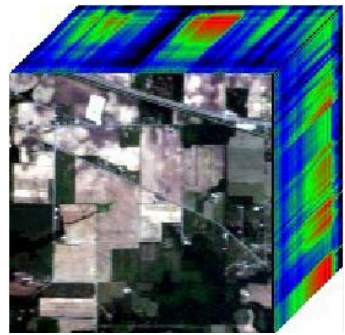

Fig 1. Hyperspectral image cube (Indian Pines)

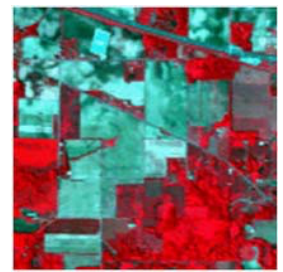

Fig 2. AVIRIS Indian Pines 3 colors image (R: 0.8314 µ m G: 0.6566 µ m B: 0.5574 µ m ).

The rest of the paper is organized as follows: part2 reviews the information theory basics, the bands selection strategies and related work. The part3 presents the new filter used to identify the most discriminative bands while part 4 outlines the experiment conducted, drawbacks of previous methods and presents the results comparison and analysis. Finally, Section 5 concludes the paper.

## II. BAND SELECTION AND RELATED WORK

Dimensionality reduction based on band selection is an essential step for hyperspectral images processing.
The selection consists on retaining the physical meaning of the dataset by selecting a set of bands from the input hyperspectral data with high discriminative power and discarding those that are irrelevant, redundant and noisy. These methods includes the sequential methods [6], the independent component analysis (ACI) represented by the work of Hyvarinen[7], and Mura [8].The principal component analysis (PCA) represented by Hwang [9], the support vector machine (SVM) classifier that proves high performance in several domains for dimensionality reduction application Guo[10]. The development of bands selection has two major directions [5].

- The wrapper methods evaluate the "goodness" of the selected bands directly based on the classification accuracy. The classifier itself is used to determine the relevance of the band. The wrapper methods perform well as the picked bands are optimized for the classification accuracy. However, these methods are very expensive in terms of computational complexity, especially when handling extremely high-dimensional data as the hyperspectral images.

- The filter methods are classifier independent. They require no feedback from classifier and estimates the classification performance indirectly by selecting the subset attributes that maximizes a certain evaluation function.

The main advantages of the filter methods in comparison with the wrappers are their computational efficiency, simplicity and independence from the classifier. Therefore, the topic of bands filter strategies that has been reviewed in detail in a number of recent articles will be the main topic of the work presented in this paper. Many filter-selection methods have been developed using different measures for bands evaluation. The evaluation function is calculated using distance, information, correlation and consistency measures. Information measure is one of the widely used measures in filtering methods.

In informative theory [11], Shannon entropy denoted by $H(A)$ is used in order to quantify the amount of information contained by a random variable A. It is defined as:

$$H(A)= \ p(A)\log 2.p(A) \qquad (1)$$

Where p(A) is the probability density function (pdf) of A.
The mutual information (MI) measures the statistical dependence between two random variables [13]. MI has been widely used as the similarity measure in hyperspectral band selection [10][12]. It is the quantity of shared information between two variables.

$$I(A,B)= \ p(A,B)\log 2.(P(A,B)/P(A).P(B)) \qquad (2)$$

Where p(A,B) denotes the joint pdf of A and B.

When applied to band selection, the mutual information is used to find the relation between a band noted (A) and the ground-truth noted (B). High Mutual Information indicates that the bands are strongly related, while zero Mutual Information shows that they are independent.
The relation between entropy and MI can be formulated as:

$$I(A,B)=H(A)-H(A/B)=H(B)-H(B/A) \qquad (3)$$

$$I(A,B)=H(A)+H(B)-H(A,B) \qquad (4)$$

Where H(A,B) is the joint entropy of two bands A and B. H(A/B) and H(B/A) are the conditional entropies of a band with the other band given.

This relation is also called information gain ("IG") the band that has the highest mutual information is considered the most informative and discriminative one. Guo [10] uses a filter based on MI to select and classify HSI AVIRIS 92AV3C [16]. Guo [10] uses also the average of bands 170 to 210, to product an estimated ground-truth map, and use it instead of the real truth map as illustrated on figure 3.

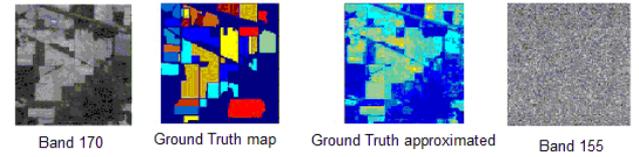

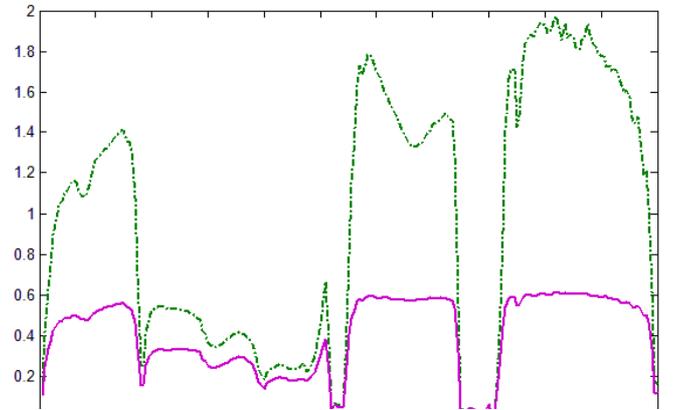

Fig 3. Calculated Mutual information of the AVIRIS with the Ground-Truth (solid line). AVIRIS with the ground approximated by averaging bands 170 to 210 (dashed line) [12].

Sarhrouni [12] uses also Mutual information to develop several schemes in maintaining the integrity of the specifications order to reduce dimensionality applied to hyperspectral images 'HSI'. The basic idea of the mutual information-based filter approach is to proceed the bands selection using a forward algorithm. The band that has the largest value of mutual information with the ground-truth is a good approximation of it. Thus, the subset of suitable bands is the one that generates the closest estimation to the GT called

GTest. We generate the current estimation by the average of the last estimation of the GT with the candidate band.
The following table presents the selection process by Sarhrouni:

| **The reproduced MI filter:** Extracted from [12] |
|---|
| *1. Ordering the bands according to the decreasing value of their mutual information with the ground-truth (GT) => **MI(GT,B)*** |
| *2. Initializing the selected bands subset using the band that has the largest mutual information value with the ground-truth (GT).* |
| *3. Using the first selected band in the previous step to build an approximation of the ground-truth denoted (**GTest**).* |
| *4. Fixing the stopping criteria: the number of retained bands chosen at the beginning of the process or when the bands are empty.* |
| *5. Introducing a threshold (Th) to control the redundancy.* |
| *6. On each iteration, we calculate the mutual information between the new (GTest) using the last added band and the (GT) =>**MI(GT,GTest).** The last added band must increase the final value of MI(GTest,GT), otherwise, it will be rejected from the choices.* |

## III. THE PROPOSED ALGORITHM

### A. Critics of previous work

In general, most of the methods listed in the previous section are based on bands selection using mutual information and consisting of two elements: the relevancy term and the redundancy term. The methods attempt to simultaneously maximize the relevancy term presented by the information gain (IG) and minimizing the redundancy term controlled by the threshold (Th). Despite their simplicity and rapidity, they represent a number of limitations.

For example, the information gain (IG) by ranking the bands based on their mutual information with the ground-truth MI(B,GT). This method assumes the independency between bands, which is not always true. IG may select relevant strongly related bands that carry redundant information about the ground-truth, which increases the computational overhead. This problem has been partly solved by Sarhrouni [12] using the redundancy term controlled by the threshold.

Yet another problem particular to the threshold. It should be manually adjusted wish creates the risk that the algorithm may lose its selection action if we choose the wrong threshold as illustrated in the result part.

Another limitation is the redundancy term that is calculated based on the value of the MI between the candidate band (new GTest) and the ground-truth (GT), without any consideration of the bands already selected. The bands may not share information with the ground-truth but that does not mean they are irrelevant. They maybe be very important for classification when combined with other bands. The previous approaches neglected the correlation, the complementarity and the information obtained when we combine the candidate band and the bands already selected within a subset. To solve the previous issues, we proposed a new method based on the Joint Mutual Information in this research.

### B. The proposed algorithm

The methods listed in the previous section attempt to optimize the relationship between relevancy and redundancy when selecting bands. However, a single band can be considered irrelevant based on its correlation with the ground GT but when added to the other bands, it brings important information to discriminate the class.

Within the informative theory research, the joint mutual information was proposed by Yang and Moody [14] using a feature selection method based on the following criteria eq (7). In existing literature Brown et al. [15] confirm also that this method performs well in terms of classification accuracy and stability.

The conditional mutual information of two random variables A and B given C is denoted as:

$$MI(A,C/B)=H(A/C)-H(A/C,B) \quad (5)$$

The joint mutual information of three random variables A, and C is denoted as JMI(A,B;C):

$$JMI(A,B;C)=MI(A,C/B)+MI(B,C) \quad (6)$$

$$JMI(A,B;C)=H(C)-H(C/A,B) \quad (7)$$

$$JMI(A,B;C)=[2:P(C)log2P(C)]-[2:log2P(AB,C/B)/(P(A/B)P(C/B)] \quad (8)$$

In the proposed method, we will use a filter approach based on the joint mutual information to ensure minimum redundancy and maximum correlation between the candidate band and the selected bands within the subset. The basic idea of the proposed forward selection approach is building the subset by adding the candidate band that maximizes the Joint Mutual Information the candidate band, the ground-truth estimated (GTest) and the ground-truth (GT). It studies relevancy and redundancy, and takes into consideration the high complementarity and interaction between the bands that are already selected within the subset presented by the ground-truth estimated (GTest) and the candidate band (B) in order to produce a good approximation of the GT.

In each iteration, the current estimation of the ground-truth is calculated by the average of the last estimation of the GT with the candidate band.

Our approach based on joint mutual information choose the most relevant bands according to the following algorithm:

*The proposed approach:*

*1. We order the bands according to the decreasing value of their mutual information with the ground truth (GT) => MI(B,GT).*

*2. We initialize the subset of the selected bands by the band (B) that has the largest mutual information value with the (GT) =>B=ArgmaxMI(s). This first selected band will be used to build an approximation of the ground truth (GT), denoted (GTest).*

*3. We fix the stopping criteria: achieving the number of the retained bands required.*

*4. During each iteration, we calculate the joint mutual information between the candidate band (B), the GT and GTest => JMI (GTest, B; GT) for all the available bands. We select the band (Bs) that maximize the joint mutual information criterion. => Bs= ArgmaxJMI(s)*

*5. After each iteration, the GTest is calculated by averaging the last estimation of the GT with the selected band (Bs) => GTest =(GTest0+Bs) /2*

*6. End the process once we achieve the stopping criteria fixed at the beginning.*

## IV. EXPERIMENT, RESULTS AND DISCUSSION

### A. Dataset & SVM Classifier

To test the performance of the proposed method, the experiments were conducted on the Hyperspectral image AVIRIS 92AV3C [16] presented previously.

This hyperspectral data sets collected by the NASA's Jet propulsion Laboratory AVIRIS instrument have been selected for experimental validation in this study.

The scene was gathered by AVIRIS over the Indian Pines test site in North-western Indiana, a mixed agricultural/forested area, early in the growing season, and consist of 145*145 pixels and 220 spectral bands in the wavelength range 0.2-2.4 µm. The AVIRIS Indian Pines data set represents a very challenging classification problem dominated by similar classes.

Discriminating among the major crops has made this scene an extensively used benchmark to validate classification accuracy of hyperspectral imaging algorithms.

The ground-truth map of the AVIRIS scene is provided, but only 10366 pixels are labeled from 1 to 16. Each label indicates one from 16 classes introduced in figure 4.

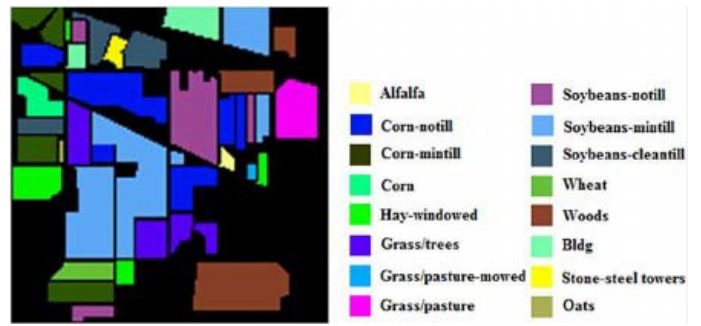

Fig 4. AVIRIS Ground truth image and its corresponding class labels.

The support vector machine classifier (SVM) [17] is used as a classifier during our tests. It is a supervised classification method based on structural risk minimization. The key idea of this technique is to estimate a separator boundary or surface between the spectral classes. This surface, which maximizes the margin between classes, uses limited numbers of boundary pixels (support vectors) to create the decision surface.

During our tests, the datasets has been divided into two parts: the training-base and the test one. Half of the pixels from each class were randomly chosen, for training, with the remaining 50% for the test set on which performance was assessed. The class labels from the ground-truth with the dataset were used for the supervised training. The average classification accuracy is used as a performance measure.

### B. Results

The following table 1 gives the classification results using the selection filter based on maximizing the mutual information (Information gain) for controlling the bands relevance.

| Number of retained Bands | The accuracy (%) of classification |
|---|---|
| 5 | 51,82 |
| 10 | 55,43 |
| 15 | 57,65 |
| 20 | 63,08 |
| 25 | 66,12 |
| 30 | 73,54 |
| 35 | 76,06 |
| 40 | 78,96 |
| 45 | 80,58 |
| 50 | 81,63 |
| 55 | 82,06 |
| 60 | 82,74 |
| 70 | 86,58 |
| 80 | 86,89 |

Table 1. Results of the information gain

The table 2 gives the reproduced classification results using Sarhrouni [12] selection approach by mutual information for different thresholds Th to control relevance and redundancy.

| | The classification accuracy for different thresholds (%) | | | | |
|---|---|---|---|---|---|
| Th | -0,02 | -0,01 | -0,0040 | -0,0035 | -0,000 |
| 2 | 47,44 | 47,44 | 47,44 | 47,44 | 47,44 |
| 3 | 47,87 | 47,87 | 47,87 | 47,87 | 48,92 |
| 4 | 49,31 | 49,31 | 49,31 | 49,31 | |
| 12 | 56,30 | 56,30 | 56,30 | 60,76 | |
| 14 | 57,00 | 57,00 | 57,00 | 61,80 | |
| 18 | 59,09 | 59,09 | 62,61 | 63,00 | |
| 20 | 63,08 | 63,08 | 63,55 | | |
| 25 | 66,12 | 64,89 | 65,38 | | |
| 35 | 76,06 | 74,72 | | | |
| 36 | 76,49 | 76,60 | | | |
| 40 | 78,96 | 79,29 | | | |
| 45 | 80,85 | 81,01 | | | |
| 50 | 81,63 | 81,12 | | | |
| 53 | 82,27 | 86,03 | | | |
| 60 | 82,74 | 85,08 | | | |
| 70 | 86,95 | | | | |
| 75 | 86,81 | | | | |
| 80 | 87,28 | | | | |

(Number of retained Bands — row labels on left)

Table 2. Results of MI algorithm [12] for different threshold

The following table 3 presents the results obtained using the proposed method based on the joint mutual information for controlling relevance, redundancy and complementarity.

| Number of retained Bands | The accuracy (%) of classification |
|---|---|
| 2 | 56,77 |
| 3 | 66,78 |
| 4 | 71,11 |
| 12 | 85,37 |
| 14 | 86,11 |
| 18 | 85,84 |
| 20 | 86,48 |
| 25 | 87,41 |
| 35 | 89,25 |
| 36 | 89,03 |
| 40 | 89,01 |
| 45 | 89,60 |
| 50 | 89,83 |
| 53 | 90,10 |
| 60 | 90,61 |
| 70 | 91,19 |
| 75 | 91,64 |
| 80 | 91,80 |

Table 3. Results of the proposed approach based on the JMI

The table 4 gives the classification results for each class using the proposed algorithm.

| Class | Total pixels | 60 bands | 70 bands | 80 bands |
|---|---|---|---|---|
| 1 | 54 | 86,96 | 86,96 | 86,96 |
| 2 | 1434 | 86,47 | 86,47 | 88,98 |
| 3 | 834 | 85,37 | 86,33 | 86,81 |
| 4 | 234 | 84,62 | 83,76 | 82,91 |
| 5 | 597 | 95,93 | 95,93 | 96,34 |
| 6 | 747 | 97,77 | 98,04 | 98,04 |
| 7 | 26 | 84,62 | 84,62 | 84,62 |
| 8 | 489 | 97,96 | 98,37 | 98,37 |
| 9 | 20 | 80,00 | 100 | 90,00 |
| 10 | 968 | 89,67 | 89,88 | 90,08 |
| 11 | 2468 | 90,28 | 91,90 | 92,14 |
| 12 | 614 | 88,60 | 91,21 | 91,53 |
| 13 | 212 | 98,06 | 98,06 | 98,06 |
| 14 | 1294 | 97,06 | 96,75 | 97,68 |
| 15 | 390 | 71,69 | 69,28 | 69,88 |
| 16 | 95 | 93,48 | 93,48 | 93,48 |

Table 4. Accuracy of classification of the proposed approach of each class

The following graph gives a comparison between algorithms based on the Average classification accuracy.

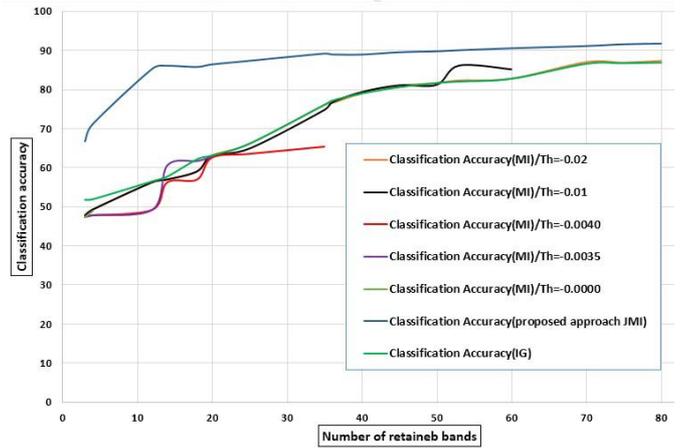

Fig 5. Average classification accuracy comparison between the algorithms

The left figure represents the reproduced map using our approach for 20 bands (86,48%), the right figure represents the ground-truth.

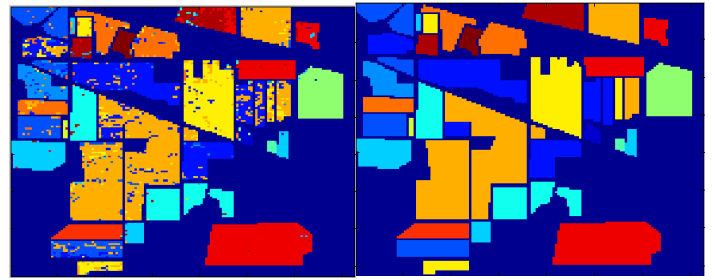

Fig 6. Original Grand Truth map ( right image) and the map produced using the proposed approach (20 bands ,86.48%) (left image)

## C. Analysis and discussion

The previous result tables (1,2,3,4) and figures (5,6) shows the classification accuracy achieved and measured using the reproduced algorithms on a subset of 2 to 80 bands.

As shown, the proposed method selects bands with high discrimination power very quickly. Our proposed method archives 86,84% classification accuracy with 20 bands, which is higher than the reproduced filter by 23%.

According to table 2, this approach uses a threshold that just be adjusted manually and correctly in order to maintain the selection action of the algorithm. It is noticeable that a few bands are selected for high threshold values (Th>-0.004) because we don't allow redundancy. For example, with Th=0, only 3 bands are retained. When we decrease the threshold (-0.01 to -0.02), some redundancy is allowed, and the classification rate increases and achieves 88% with Th=-0.02 and 83 selected bands.

Based on the figure 5, the proposed algorithm resolves the selection action issue caused by choosing inappropriate threshold at the beginning of the process. The redundancy and the interaction with the already selected band relative to the ground-truth are controlled by the joint mutual information without threshold using.

According to table 3 and the graph in figure 5, we notice that the proposed approach produced its best accuracy of 91.80% for 80 bands, which is better than the information gain filter by 4.91 % and better than the mutual information-based filter for (Th=-0,02) about 4.51%. For the number of retained bands upper than 80, the proposed algorithm reaches its maximum accuracy value equal to 91,80 %.

We can conclude that the proposed process uses a simple criterion based on the joint mutual information without including a threshold that could limit the selection process as shown in figure 5. The simplicity and quick discrimination are the most important improvements on our approach based on the simultaneous relevance, redundancy and correlation control using the joint mutual information.

## V. Conclusion

This paper presents a new band selection method based on information theory: the Joint Mutual Information criterion maximization. This method is designed to resolve the high dimensionality constrain of the hyperspectral images. More precisely to solve the problem of choosing redundant and irrelevant band in filter approach and take into consideration the complementarity between the bands to discriminate the ground-truth.

The method has been evaluated using the HSI AVIRIS 92AV3C and the support vector machine classifier and compared with two reproduced filters methods based on mutual information.

The results shows that the proposed algorithm based on the joint mutual information improves the classification accuracy with a lesser number of bands compared to the reproduced approaches. The results demonstrate the ability of the proposed method to select a bands subset with high discriminative power and confirm that the analysis of the joint mutual information between the bands must be taken into consideration during the selection process.

## References

[1] Plaza, Antonio, et al. "Recent advances in techniques for hyperspectral image processing." Remote sensing of environment 113 (2009): S110-S122.

[2] Fauvel, Mathieu, et al. "Advances in spectral-spatial classification of hyperspectral images." Proceedings of the IEEE 101.3 (2013): 652-675.

[3] Hughes, Gordon P. "On the mean accuracy of statistical pattern recognizers." Information Theory, IEEE Transactions on 14.1 (1968): 55-63.

[4] Jain, Anil K., Robert PW Duin, and Jianchang Mao. "Statistical pattern recognition: A review." Pattern Analysis and Machine Intelligence, IEEE Transactions on 22.1 (2000): 4-37.

[5] Sarhrouni, Elkebir, Ahmed Hammouch, and Driss Aboutajdine. "A dashboard to analysis and synthesis of dimensionality reduction methods in remote sensing." Int. J. Eng. Technol 5.3 (2013): 2678-2684.

[6] A. Jain et D. Zongker. Feature selection: evaluation, application,and small sample performance. IEEE Trans. Pattern Anal. Machine Intell,vol. 19, pages 153-158, 1997.

[7] A. Hyvarinen et E. Oja. Independent component analysis: algorithms and applications. Neural Netw., vol. 13, no. 4-5, pages 411-430, 2000.

[8] M. D. Mura, A.Villa, J. A. Benediktsson, J. Chanussot et L. Bruzzone. Classification of Hyperspectral Images by Using Extended Morphological Attribute Proles and Independent Component Analysis. IEEE GEOSCIENCE AND REMOTE SENSING LETTERS, VOL. 8, NO. 3, MAY 2011, vol. 8, no. 3, pages 541-545, 2011.

[9] J. Hwang, S. Chen et J. Ji. A Study of Hyperspectral Image Classification Based on Support Vector Machine. Proceedings of Asian Conference on Remote Sensing (ACRS) ACRS2009, 2009.

[10] B. Guo, Steve R. Gunn, R. I. Damper Senio et J. D. B. Nelson. Customizing Kernel Functions for SVM-Based Hyperspectral Image Classification. IEEE TRANSACTIONS ON IMAGE PROCESSING, vol. Vol 17, no. 4, pages 622-629, 2008.

[11] Cover, Thomas M., and Joy A. Thomas. "Elements of information." (1991).

[12] E. Sarhrouni, A. Hammouch et D. Aboutajdine. Dimensionality reduction and classification feature using mutual information applied to hyperspectral images: a filter strategy-based algorithm. Appl. Math. Sci, vol. 6, no. 101-104, pages 5085{5095, 2012. (Cited in pages 2, 24, 25,29, 34, 35, 36, 40 et 49.).

[13] Kullback, Solomon. Information theory and statistics. Courier Corporation, 1968.

[14] Yang, H., & Moody, J. (1999). Feature selection based on joint mutual information. In Proceedings of international ICSC symposium on advances in intelligent data analysis (pp. 22–25).

[15] Brown, G., Pocock, A., Zhao, M., & Lujan, M. (2012). Conditional likelihood maximisation: a unifying framework for information theoretic feature selection. Journal of Machine Learning Research, 13, 27–66.

[16] D. Landgrebe, On information extraction principles for hyperspectral data: A white paper, Purdue University, West Lafayette, IN, Technical Report, School of Electrical and Computer Engineering, 1997.

[17] Chih-Chung Chang and Chih-Jen Lin, LIBSVM: a library for support vector machines. ACM Transactions on Intelligent Systems and Technology, 2:27:1- 27:27, 2011. Software available at http://www.csie.ntu.edu.tw/ cjlin/libsvm.